\def\year{2020}\relax
\pdfoutput=1
\documentclass[letterpaper]{article} 
\usepackage{aaai19}  
\usepackage{times}  
\usepackage{helvet} 
\usepackage{courier}  
\usepackage[hyphens]{url}  
\usepackage{graphicx} 
\usepackage{graphicx}  
\usepackage{xcolor}
\usepackage{amsmath}
\usepackage{url}
\usepackage{balance}
\usepackage{color}
\usepackage{times}
\usepackage{graphicx}
\usepackage{multirow}
\usepackage{multicol}
\usepackage{latexsym}
\usepackage{url}

\def\vs{\vspace{-2pt}}

\definecolor{darkgreen}{rgb}{0.13, 0.55, 0.13}
\title{A Lexical, Syntactic, and Semantic Perspective for Understanding Style in Text}
\author{Gaurav Verma, Balaji Vasan Srinivasan\\
BigData Experience Lab\\
Adobe Research, India\\
\texttt\{gaverma, balsrini\}@adobe.com
}
\begin{document}
\maketitle
\begin{abstract}
With a growing interest in modeling inherent subjectivity in natural language, we present a linguistically-motivated process to understand and analyze the writing style of individuals from three perspectives: lexical, syntactic, and semantic. We discuss the stylistically expressive elements within each of these levels and use existing methods to quantify the linguistic intuitions related to some of these elements. We show that such a multi-level analysis is useful for developing a well-knit understanding of style -- which is independent of the natural language task at hand, and also demonstrate its value in solving three downstream tasks: authors' style analysis, authorship attribution, and emotion prediction. We conduct experiments on a variety of datasets, comprising texts from social networking sites, user reviews, legal documents, literary books, and newswire. The results on the aforementioned tasks and datasets illustrate that such a multi-level understanding of style, which has been largely ignored in recent works, models style-related subjectivity in text and can be leveraged to improve performance on multiple downstream tasks both qualitatively and quantitatively.

\end{abstract}

\section{Introduction}
\label{sec:introduction}

Modeling inherent subjectivity in natural language is of key importance for making advances in computational social science.  While the notions of subjectivity pertaining to sentiment and opinion have attracted attention from computational linguists, a similar linguistic analysis of writing style has been missing from the current literature. Nonetheless, there has been a growing interest in \textit{solving} \textit{tasks} related to style in text \cite{W18-1505,niu2018polite,fu2017style}. These approaches, however, have been limited due to their assumptions about style and its composition. They use stylistic intuitions that are linked to \textit{differences} in style -- be it genre classification \cite{kessler1997automatic}, author profiling \cite{garera2009modeling}, social relationship classification \cite{peterson2011email}, readability classification \cite{collins2005predicting}, stylized text generation \cite{hovy1990pragmatics,inkpen2006building} or style transfer \cite{li2018delete,prabhumoye2018style}.  
These assumptions are often task-specific and do not cover all aspects of style leading to a need to fill the gap between the understanding of style and solving tasks related to it. In this work, we present a linguistically-motivated process to develop a task-independent understanding of style -- that is not tied to any of the above tasks, and is general enough to encompass them. 

Stylistic variations in language are reflections of factors like context, author-reader dynamics, and the backgrounds of the parties involved. The influence of these factors has been analyzed in detail by psycholinguists \cite{semino2002cognitive,enkvist1985introduction}. Linguistic style also deals with the prescriptive grammar associated with the aesthetics of text, as analyzed by computational linguists \cite{lakoff1979stylistic,thurmair1990parsing}. Our exploration in this paper is centered around the computational linguistics perspective.

Earlier efforts of understanding style in text \cite{strunk1959elements,dimarcoStyle,crystal2016investigating} focus on laying out \textit{stylistic elements}, defined as the components of language that are stylistically expressive. Using this definition, we identify and discuss the elements at \textbf{\textit{lexical}} (relating to the  vocabulary of a language), \textbf{\textit{syntactic}} (relating to the arrangement of words and phrases to create well-formed sentences) and \textbf{\textit{semantic}} (relating to meaning in language) levels. Our qualitative analysis highlights key aspects of style that have been ignored in recent works \cite{W18-1505,prabhumoye2018style,jhamtani2017shakespearizing}. We follow our qualitative analysis with a discussion of existing methods that can be used to quantify these aspects and facilitate their computational modeling. We demonstrate the value of a well-knit understanding of style by solving three downstream tasks: \textbf{\textit{analysis of writing style}} of $5$ popular English authors, \textbf{\textit{authorship attribution}} and \textbf{\textit{emotion prediction}}. 
We conduct experiments on datasets that cover a diverse range of topics and domain, comprising of social media posts (Facebook and Twiiter), user reviews (IMDb), legal documents, literary books, and newswire articles (Reuters). 

The rest of the paper is structured as follows: we start by discussing the proposed multi-level structure to understand style in text with several qualitative examples. In the subsequent section, we quantify representative stylistic elements to demonstrate the value of such an understanding in analyzing authors' style. In the following sections, we further use these quantifications to solve the tasks of authorship attribution and emotion prediction. We discuss various insights from our results as well as the related prior work towards the end of the paper. In the final section, we conclude the work while laying out the scope for future work.
 
\section{Elements of Style in Text}
\label{sec:StyleElements}

Stylistically expressive elements in text can be identified at word-level (lexical), in the way sentences are structured (syntactic), and by analyzing the attributes of core-meaning that is conveyed (semantic). However, it must be noted that a style element can belong to one or more of the above categories \cite{dimarcoStyle}. Here, we briefly describe each of the style elements and also provide examples to demonstrate the non-trivial entanglement of style and meaning in text.

\subsubsection{Lexical Elements }of style are expressed at word-level, and the stylistic variation can arise due to addition, deletion, or substitution of words. These variations can give rise to text that is characteristically different in terms of sentiment, formality, excitement etc. For example, words like \textit{residence} and \textit{occupied} are objective in nature, and emotionally-distant from their subjective counterparts, \textit{home} and \textit{busy} \cite{brooke2013hybrid}.

We also observe change in meaning and sentiment with some word-level variations: \textit{Great food but horrible staff} vs. \textit{Great food and awesome staff} \cite{li2018delete}. 

\citeauthor{brooke2010automatic} (\citeyear{brooke2010automatic}, \citeyear{brooke2013multi}) enumerate such stylistic dimensions represented in lexicon as: colloquial vs. literary; concrete vs. abstract; subjective vs. objective; and formal vs. informal. For instance, while the words \textit{residence} and \textit{occupied} are objective, \textit{home} and \textit{busy} are subjective; while the word \textit{tasty} is colloquial, \textit{palatable} is literary.

\subsubsection{Syntactic Elements} of style are prominent in language -- some examples being, \textit{piled-up adjectives}, \textit{detached adjectival clause}, \textit{adjectival phrase}\footnote{\scriptsize{\textbf{Piled-up adjectives}: The \textit{gigantic green} dragon felt bad. \textbf{Detached adjectival clause}: He, \textit{being a recluse}, often quietly excused himself. \textbf{Adjectival phrase:} The movie was \textit{not too terrible}}. }\cite{dimarcoStyle}.

The use of active voice is more direct and energetic than passive \cite{strunk1959elements}. Rhetorical theories have contrasted \textit{loose} and \textit{periodic} sentences -- placing the most important clause at the beginning vs. placing it at the end of a sentence\footnote{\scriptsize{\textbf{Loose sentence}: ``\textit{These algorithms can not deal with words for which classifiers have not been trained}"; \textbf{Periodic sentence}: ``\textit{For processing free texts hand-crafted grammars are neither practical nor reliable.}" \cite{feng2012characterizing}}}. 
Such stylistic variations are well captured in the parse trees generated using probabilistic context-free grammar (PCFG) \cite{booth1973applying}. For instance, the parse trees of loose sentences are deeper and unbalanced, while those for periodic sentences are relatively more balanced and wider \cite{feng2012characterizing}. 

It is noteworthy that some of these syntactic style elements express themselves over multiple sentences, and are not constrained within a single sentence. For example, the use of several loose sentences in succession leads to triteness due to mechanical symmetry and a singsong effect \cite{strunk1959elements}. The classification of sentences as simple, compound, complex, and complex-compound and computing their statistics has facilitated identifying and differentiating between various author's styles \cite{feng2012syntactic}.

\subsubsection{Semantic Elements} of style can be identified by analyzing the attributes of underlying meaning that is being conveyed in a piece of text. For example, consider the two sentences: \textit{He was not very often on time} vs. \textit{He usually came late}. While both the sentences have a similar core-meaning, the former seems rather hesitating and noncommital, while the latter stands strong and resolute -- being able to express a negative in a positive form. 

Following two sentences put to contrast the vagueness and concreteness of meaning that is being conveyed: \textit{He showed satisfaction as he took possession of his well-earned reward} vs. \textit{He grinned as he pocketed the award} \cite{strunk1959elements}. It should be noted that semantic elements of style are identified by analyzing the larger meaning of the text (phrase, sentence, or paragraph), unlike lexical or syntactic elements which consider the meaning of the comprising words or the syntax of the underlying sentence.

Various stylistically expressive elements can occur in conjunction.  \citeauthor{dimarcoStyle} present an interesting example where a detached adjectival clause, which is a syntactic element of style, can present variations that are semantically concordant and discordant: \textit{The university attended  by the President, one of the finest law schools in the country, is the alma mater of many politicians} vs. \textit{The university attended  by the President, a set of building with an architectural charm of a prison, is the alma mater of many politicians}.

Apart from these $3$ elements of style, there are \textbf{surface-level style elements} that relate to length of sentences, use of punctuation, length of words, length of paragraphs, etc. For instance, use of Oxford commas is mandated by the American Psychological \citeauthor{american2010concise} but not recommended by the Associated Press \cite{goldstein1998associated}.

\begin{table*}[!h]
\centering\vs\vs
\scalebox{0.80}{
\begin{tabular}{l|l| c | c | c | c |c }
  \textbf{Level} & \textbf{Stylistic Element} & \textbf{Abraham Lincoln} & \textbf{Mark Twain} & \textbf{Oscar Wilde} & \textbf{Rudyard Kipling} & \textbf{Charles Dickens}\\
  \hline
  \multirow{3}{*}{\textbf{\textit{Surface}}} & Avg \# of words in a sentence $(\mu \pm \sigma)$  & \textbf{30.88} $\pm$ $4.16$ & \textbf{23.22} $\pm$ $2.64$ & 20.35 $\pm$ $2.91$ & \textbf{19.29} $\pm$ $3.12$ & \textbf{26.77} $\pm$ $3.91$ \\
     & Avg \# of commas; semicolons; colons  &\textbf{1.28}; 0.11; 0.03  & 1.14; 0.31; 0.08 & 0.82; 0.05; 0.01 & 0.93; 0.57; 0.02  & \textbf{1.42}; 0.19; 0.03 \\
     & Avg \# of sentences in a para. $(\mu \pm \sigma)$  & \textbf{3.52} $\pm$ $0.63$ & 5.59 $\pm$ $0.42$ & 5.19 $\pm$ $0.39$ & 5.35 $\pm$ $0.47$ & 4.53 $\pm$ $0.57$ \\\hline
  \multirow{4}{*}{\textbf{\textit{Lexical}}} & Literary vs. colloquial words & \textbf{0.91} vs. 0.09 & 0.73 vs. 0.27 & 0.69 vs. 0.31 & 0.61 vs. 0.39 & 0.75 vs. 0.25 \\
   & Abstract vs. concrete words & \textbf{0.71} vs. 0.29 & 0.64 vs. 0.36 & 0.58 vs. 0.42 & \textbf{0.39} vs. \textbf{0.61} & 0.54 vs. 0.46 \\
   & Subjective vs. objective words & 0.57 vs. 0.43 & 0.58 vs. 0.42 & 0.68 vs. 0.32 & \textbf{0.48} vs. \textbf{0.52} & 0.57 vs. 0.43 \\
   & Formal vs. informal words &\textbf{ 0.87} vs. 0.13 & 0.74 vs. 0.26 & 0.70 vs. 0.30 & 0.57 vs. 0.43 & 0.62 vs. 0.38 \\\hline
  \multirow{4}{*}{\textbf{\textit{Syntactic}}} & Fraction of simple sent. & \textbf{0.09} & \textbf{0.18} & 0.17& \textbf{0.26}  & \textbf{0.13} \\
   & Fraction of compound sent. & 0.21  & 0.23  & 0.24  & 0.24  & 0.21  \\
   & Fraction of complex sent. & \textbf{0.31}  & 0.27  & 0.26  & 0.25  & \textbf{0.30}  \\
   & Fraction of complex-compound sent. & \textbf{0.35}  & 0.28  & 0.26  & 0.23  & \textbf{0.29} \\
   & Fraction of loose sentences & \textbf{0.17}  & 0.09 & 0.07 & \textbf{0.03} & 0.16 \\
   & Fraction of periodic sentences & \textbf{0.12}  & 0.14 & 0.11 & \textbf{0.08} & 0.12 \\

\end{tabular}\vs\vs
}
\caption{{\small{\textit{Quantification of stylistic elements at different levels, for author analysis.}}} \small{{The presented values are averages computed over all $10$ documents for each author. For surface-level analysis, we report the averages and standard deviations ($\mu \pm \sigma$). For lexical analysis, the values for $a$ vs. $b$ are equal to} $\frac{\text{\# of words in $a$}}{\text{\# of words in $a$ or $b$}}$ 
{and} $ 1- \frac{\text{\# of words in $a$}}{\text{\# of words in $a$ or $b$}}$ {respectively. For syntactic analysis, the sum of all fractions corresponding to simple, compound, complex, and complex-compound is $\leq 1$, owing to sentences that do not fall into either of those categories -- \textit{What an idiot!} being one such example of an incomplete sentence. The values in \textbf{bold} are of interest and have been mentioned in the text.}} \vs\vs\vs\vs\vs\vs\vs
}
\label{tab:results}
\end{table*}

\subsubsection{Separability of Meaning and Style:} Let us consider the following set of sentences:\\
$S1$: Chocolates can kill you.\\
$S2$: Chocolates, although tasty, can kill you.\\
$S3$: Chocolates, although palatable, can kill you.\\
$S4$: A period of unfavorable weather set in.\\
$S5$: It rained every day for a week.\\
From our discussion above, $S2$ and $S3$ present variation at lexical-level -- $S3$ is literary while $S2$ is more colloquial. $S1$ and $S2$ express variation at syntactic-level -- $S2$ has an additional detached adjective. $S4$ and $S5$ illustrate variation at semantic-level -- one can notice that the core-meaning in $S4$ is vague while $S5$ is definite and specific.

These examples illustrate that style is not merely a means of expressing meaning, but plays an active role in bringing about meaning. While there are variations in style that are separable from meaning (for example, the syntactic switch in style between active and passive voice: \textit{I shall always remember my first visit to Boston} vs. \textit{My first visit to Boston will always be remembered}), the disentanglement of style and meaning can only be applied to a few stylistic elements and the process becomes increasingly complex as we progress from lexical, to syntactic, to semantic variations.

Recent approaches in style-related tasks assume independence between meaning and style \cite{li2018delete,jhamtani2017shakespearizing,prabhumoye2018style} restricting the understanding of style to a means of reflecting already existing meaning \cite{tikhonov2018wrong,dai2019style}. For example, recent style transfer approaches involve generating realization $\mathbf{x}_2$ tuned to style $\mathbf{y}_2$, from realization $\mathbf{x}_1$ possessing style $\mathbf{y}_1$, by learning an auto-encoder model, that infers $\mathbf{x}_1$'s latent meaning  $\mathbf{z} \sim p(\mathbf{z} \vert \mathbf{x}_1, \mathbf{y}_1)$ and then generates the transferred counterpart from $p(\mathbf{x}_2 \vert \mathbf{z}, \mathbf{y}_2)$ \cite{shen2017style}. This approach is useful where style $\mathbf{y}_1$ can be separated from $\mathbf{x}_1$ to represent the meaning $\mathbf{z}$, \textit{and} the \textit{same} meaning can be used to generate realization $\mathbf{x}_2$ tuned to $\mathbf{y_2}$. While their approach can handle most of the lexical-level  variations (see $S2$ vs. $S3$), they cannot handle other variations at syntactic (see $S1$ vs. $S2$) or semantic-level ($S4$ vs. $S5$). Recently, registering these shortcomings of existing work, \citeauthor{dai2019style} (\citeyear{dai2019style}) aim to perform the task of style transfer without learning disentangled latent representations. Our work is a step towards understanding linguistic aspects of style and can be used to expand on the approaches to solve style-related natural language tasks.

\section{Authors' Style Analysis}
\label{sec:Quantitative}
We quantify a few representative stylistic elements\footnote{We do not aim to quantify \textit{all} the discussed stylistic elements. The primary goal is to provide empirical evidence to substantiate the claim that such a multi-level understanding of style is of value in solving downstream tasks. Quantification of some of the discussed stylistic elements, especially the ones at semantic-level, is an open problem in itself and calls for further research.} at each of these levels and analyze the writings of 5 popular English authors (Abraham Lincoln, Mark Twain, Oscar Wilde, Rudyard Kipling and Charles Dickens) from the Gutenberg dataset \cite{lahiri:2014:SRW}.
The subset of Gutenberg dataset that we consider for analyzing the writing style of authors comprises of $50$ published books ($10$ books per author). Each of the books contain, on average, $\sim1,000$ sentences and the entire corpus of $50$ books contains around $25,000$ unique words (such that their frequency of occurrence is greater than or equal to $3$ in the entire corpus of $50$ books). In the supplementary document, for the ease of reproducibility, we provide the author-wise list of books and \textit{representative} writing samples from the five authors under consideration.

At surface-level, we report the average number of words in a sentence, average number of commas, semicolons and colons in a sentence, and average number of sentences in a paragraph.  For quantifying the lexical elements, we use a list of seed words for each of the following eight categories: subjective, objective, concrete, abstract, literary, colloquial, formal and informal \cite{brooke2013multi}. Following \citeauthor{brooke2013multi} (\citeyear{brooke2013multi}), we compute normalized  pointwise  mutual  information  index  (PMI)  to  obtain  a  raw  style score for each dimension,  by  leveraging  co-occurrences  of words in the entire corpus. The raw scores are  normalized to  obtain  style  vectors  for  every  word,  followed by a transformation of style vectors into k-Nearest Neighbor (kNN) graphs, where label propagation is applied. It is worth noting that the eight original dimensions lie on the two extremes of four different spectrums, i.e., subjective-objective, concrete-abstract, literary-colloquial, and formal-informal. We compute averages across a given author's corpus. The averages, in the range $[0, 1]$, denote the author's tendency to use subjective, concrete, literary, or formal words, in contrast to using objective, abstract, colloquial, or informal words, respectively, as evidenced in their literary works.

For syntactic analysis, we use sentence structure identification algorithms proposed by \citeauthor{feng2012characterizing} (\citeyear{feng2012characterizing}), and compute the fractions of simple, compound, complex, and complex-compound sentences. The numbers reported in Table \ref{tab:results} are fractions of all sentences in the concerned author's corpus. We also quantify the fraction of sentences that are identified as periodic and loose.

For semantic analysis, in absence of clear approaches for quantification, we observe the output of a knowledge parser, \textit{K-Parser} \cite{sharma2015identifying}, and report most frequent entities and their semantic roles \cite{palmer2005proposition}.

The statistics in Table \ref{tab:results} indicate that the quantification aligns with several qualitative observations pertaining to the authors. For instance, writings of Abraham Lincoln involve statements of political significance and hence, are carefully structured (see the supplementary material for representative writing samples). At surface-level, this is reflected in longer sentences, extensive use of commas, and fewer sentences in a paragraph. At lexical-level, the use of abstract words like \textit{freedom}, \textit{respect}, \textit{passion} stands out. It is also notable that since loose and periodic sentences typically have a non-simple syntactic structure, Lincoln has a significantly larger fraction of sentences that are either identified as loose or periodic sentences. The use of words that are more literary than colloquial and more formal than informal, is also quite prominent. At syntactic-level, Lincoln's sentences are structured complicatedly and very few sentences have a simple syntactic structure. Lincoln often referred to a nation as a living entity, which is in turn observed as \textit{nation} being one of the frequently used entities with its semantic role being \texttt{\small{\textit{:alive entity}}}. Along similar lines, the semantic role for the entity \textit{law} was frequently found to be \texttt{\small{\textit{:impelling agent}}}.

Mark Twain, while having similar lexical-level polarities as Lincoln, uses relatively simpler sentences. Oscar Wilde, known for his satire on contemporary culture \cite{ellmann1988oscar}, frequently uses the entity \textit{people} in his writings, where the two frequently associated semantic roles are \texttt{\small{\textit{:object of affection}}} and \texttt{\small{\textit{:object of disaffection}}}. It is interesting to note that the characteristically long sentences that are attributed to Charles Dickens \cite{hobsbaum1998reader} are also captured in this multi-level analysis of style -- significantly higher number of words in a sentence, with prominent use of punctuation for conjunctions (i.e., commas), and a higher tendency to use complex and complex-compound sentences.

The stylistic \textit{variations} in the writings are well captured across all the levels. For example, Rudyard Kipling, well-known for short stories and classics of children's literature \cite{angus1979strange}, has a higher tendency of forming short sentences with simple syntactic structure than other authors -- which is in turn also reflected in comparatively lower fractions of loose/periodic sentences. Additionally, Kipling is the only author to use more concrete words like \textit{gongs}, \textit{rockets}, \textit{torch} etc. and less abstract words like \textit{suffer}, \textit{freedom}, etc.

These observations reinforce that writing style is a compound factor of several stylistic elements and can be identified at multiple levels in text, validating the need of a multi-level analysis. The \textit{differences} in writing style of authors that are observed across these levels further strengthen the support for such a multi-level analysis.

\section{Authorship Attribution}
\label{sec:authorAttribution}

\begin{table}[!t]
\centering\vs\vs
\scalebox{0.75}{
\begin{tabular}{l | c | c | c | c  }
  \textbf{Characteristics} & {\textbf{Judgement}} & {\textbf{CCAT10}} & {\textbf{CCAT50}} &{\textbf{IMDb62}}\\
  \hline
  genre & legal & \multicolumn{2}{c|}{newswire} & movie reviews\\
  \# authors & $3$ & $10$ & $50$ & $62$\\
  \# of documents & $1,342$ & $1000$ & $5000$ & $79,550$\\
  avg chars / doc & $11,957$ & $3,089$ & $3,058$ & $1,401$\\
  avg words / doc & $2,367$ & $580$ & $584$ & $288$\\
\end{tabular}\vs\vs\vs\vs
}
\caption{\small{Data statistics for the authorship attribution task}}
\label{tab:data_stats}
\end{table}

\begin{table*}[thb]
\centering\vs\vs
\scalebox{0.72}{
\begin{tabular}{l | c  | c | c | c }
  {\textbf{Models}} & {\textbf{Judgement}} & {\textbf{CCAT10}} & {\textbf{CCAT50}} & {\textbf{IMDb62}}\\
  \hline
  {SVM with bag of local histogram \cite{escalante2011local}} & -- & {$\mathbf{86.40}\%$} & -- & -- \\\hline
  Token SVM \cite{seroussi2014authorship} & $91.15\%$ & -- & -- & $91.52\%$ \\\hline
  Using topic models \cite{seroussi2014authorship} & {$93.64\%$} & -- & -- & {$91.79\%$} \\\hline
  Document embeddings based on textual style \cite{parikhtowards} & -- & $63.80\%$ & $76.60\%$ & $89.90\%$\\\hline
  $^\dagger$Continuous character n-grams + content \& style \cite{sari2017continuous,sari2018topic} & {$91.51\%$} & $76.20\%$ & {$72.88\%$} & {$95.93\%$} \\\hline
  Continuous character n-grams + content \& style + \textbf{New features}  & {$\mathbf{94.44}\%$} & {$80.07\%$} & {$\mathbf{77.75}\%$} & {$\mathbf{97.89}\%$} \\\hline
\end{tabular}\vs\vs\vs\vs
}
\caption{{\small{\textit{Performance of our proposed approach on the authorship attribution task}.}} \small{We concatenate multi-level stylistic features with the auxiliary features of the baseline$^\dagger$ and compare the classification accuracies. }}\vs\vs\vs\vs 
\label{tab:results_AA_SoTA}
\end{table*}

\begin{table}[thb]
\centering\vs\vs
\scalebox{0.75}{
\begin{tabular}{l | c | c | c | c }
  {\textbf{Features}} & {\textbf{Judgement}} & {\textbf{CCAT10}} & {\textbf{CCAT50}} & {\textbf{IMDb62}}\\\hline
  baseline features$^\dagger$ & {$91.51\%$} & {$76.20\%$} & {$72.88\%$} & {$95.93\%$} \\\hline
  (+) lexical &$\underline{+1.17}$& $\underline{+2.17}$  & $\underline{+2.87}$ & $\underline{+0.93}$ \\
  (+) surface  & $+0.09^*$ & $\underline{+0.13}$ &  $\underline{+0.07}$ &  $\underline{+0.11}$ \\
  (+) syntactic & $\underline{+1.57}$ & $\underline{+1.29}$  & $\underline{+1.21}$ &  $+0.87^*$ \\
  (+) all  & $\underline{+2.91}$ & $\underline{+3.87}$ &  $\underline{+4.87}$ & $\underline{+1.96}$ \\
\end{tabular}\vs\vs\vs\vs
}
\caption{{\small{\textit{Stylistic feature ablation results for authorship attribution task}.}} \small{$+$ denotes \% increase over baseline$^\dagger$ due to addition of proposed stylistic features. Underlined values are statistically significant with $p < 0.001$, while those with $^*$} are significant with $p < 0.01$.}\vs\vs\vs\vs\vs\vs\vs
\label{tab:results_AA_Ablation}
\end{table}

\begin{table*}[bth]
\centering\vs\vs
\scalebox{0.80}{
\begin{tabular}{l | c  c  c | c  c }
  \multirow{2}{*}{\textbf{Models}} & \multicolumn{3}{c|}{\textbf{EmoBank}} & \multicolumn{2}{c}{\textbf{FB Post} }\\
  {} & Val & Aro & Dom & Val & Aro \\
  \hline
  {System \cite{preoctiuc2016modelling}}& {--} & {--} & {--} & {$0.650$} & {$\mathbf{0.850}$}\\\hline
  {CNN (C) \cite{akhtar2018multi}}& {$0.567$} & {$0.347$} & {$0.234$} & {$0.678$} & {$0.290$}\\
  {LSTM (L)} & {$0.601$} & {$0.337$} & {$0.245$} & {$0.671$} & {$0.324$}\\
  {GRU (G)} & {$0.569$} & {$0.315$} & {$0.243$} & {$0.668$} & {$0.313$}\\
  {Ensemble (CLG)} & {$0.618$} & {$0.365$} & {$0.263$} & {$0.695$} & {$0.336$}\\
  {$^\ddagger$Ensemble (CLG + Old features) } & {$0.635$} & {$0.375$} & {$0.277$} & {$0.727$} & {$0.355$}\\\hline
    \hspace{60pt}(+) lexical & {$\underline{+0.026}$} & {$\underline{+0.003}$} & {${+0.017^*}$} & {$\underline{+0.010}$} &{ $\underline{+0.043}$} \\
  \hspace{60pt}(+) surface  & {${-0.003^*}$} & {$\underline{-0.005}$} & {$\underline{-0.001}$} & {$-0.006^*$} & {$\underline{+0.008}$}\\
  \hspace{60pt}(+) syntactic & {$\underline{+0.012}$} & {$\underline{+0.006}$} & {${+0.002^*}$} & {${+0.004^*}$} & {$\underline{+0.009}$} \\
  \hspace{60pt}(+) all  & {$\underline{+0.039}$} & {$\underline{+0.007}$} & {${+0.020^*}$} & {$\underline{+0.009}$} & {$\underline{+0.062}$} \\
  {Ensemble (CLG + Old \& \textbf{New features})} & {$\mathbf{0.674}$} & {$
  \mathbf{0.382}$} & {$\mathbf{0.297}$} & $\mathbf{0.736}$ & $0.417$\\\hline
\end{tabular}\vs\vs\vs\vs}
\caption{\small{\textit{Performance on the emotion prediction task.} We concatenate multi-level stylistic features with handcrafted features and compare Pearson coefficient correlation. $+$ denotes increase over baseline$^\ddagger$ due to addition of proposed stylistic features. Underlined values are statistically significant with $p < 0.001$, while those with $^*$ are significant with $p < 0.01$.} }\vs\vs\vs\vs 
\label{tab:results_EP_SoTA}
\end{table*}

To further illustrate the value of a task-independent understanding of style, we use the quantified style elements to solve the authorship attribution task -- the task of identifying the author of a document \cite{love2002attributing}. We use the methods by \citeauthor{sari2018topic} (\citeyear{sari2017continuous,sari2018topic}) as the baseline, and analyze the effects of adding our multi-level stylistic features. We use four datasets --  Judgement \cite{judgement}, CCAT10, CCAT50 \cite{stamatatos2008author}, and IMDb62 \cite{seroussi2010collaborative} which cover a range of characteristics in terms of number of authors, topic, and document length \cite{sari2018topic} (see Table \ref{tab:data_stats} for details). We concatenate our quantified stylistic features at lexical, surface and syntactic level ($4$, $3$, and $6$ in number, respectively; see Table \ref{tab:results}) with the features 
designed by \citeauthor{sari2018topic} (\citeyear{sari2018topic}). The original features that \citeauthor{sari2018topic} use are designed to capture authors' writing style and topical preference. 

\subsubsection{Baselines:} \citeauthor{sari2017continuous} (\citeyear{sari2017continuous}) propose  to represent a document as a bag of character-based n-gram features and learn the continuous representation of each feature jointly with the classifier in a shallow
feed-forward neural network. Following this work, in \citeyear{sari2018topic} they extend their character-based model by incorporating a combination of style and content related features as auxiliary features represented in discrete form. Their style-based features include features like average word length, number of short
words, frequency of function words, occurrence of punctuation, etc., while their content-based features include frequency of uni/bi/tri-grams of common words. These auxiliary features provide additional information related to the dataset characteristics. We concatenate our proposed multi-level stylistic features to these auxiliary features and analyze their efficacy. It is worth noting that to isolate the effects of modeling changes and input feature changes, we keep the hyperparameters same as those in the baseline models.
We also compare the performance of our proposed approaches with other state-of-the-art models \cite{seroussi2014authorship,escalante2011local,parikhtowards}.
For comparisons in Table \ref{tab:results_AA_SoTA} and \ref{tab:results_AA_Ablation}, we use the same data preprocessing techniques and model hyperparameters as described in their work by \citeauthor{sari2018topic} (\citeyear{sari2017continuous,sari2018topic})\footnote{\scriptsize{URL for reproducing the baselines: \textit{https://github.com/yunitata/continuous-n-gram-AA} and \textit{https://github.com/yunitata/coling2018}}}. Please refer to \citeauthor{sari2017continuous} (\citeyear{sari2017continuous,sari2018topic}) for further details.
 
Table \ref{tab:results_AA_SoTA} summarizes the effect of using proposed stylistic features along with existing features with respect to the continuous character n-grams model \cite{sari2017continuous}. In Table \ref{tab:results_AA_SoTA}, we report the average accuracy over $20$ different experimental runs. Additionally, in Table \ref{tab:results_AA_Ablation} we report the change in accuracy brought by adding \textit{(a)} new stylistic features at individual levels and \textit{(b)} the stylistic levels across \textit{all} levels. We also indicate the statistical significance (\textit{t-test}) of presented results in Table \ref{tab:results_AA_Ablation}. 

It can be noted from Table \ref{tab:results_AA_SoTA} that the inclusion of proposed multi-level stylistic features improves the performance of existing state-of-the-art models on Judgement, CCAT50, and IMDb62. More importantly, a further analysis in Table \ref{tab:results_AA_Ablation} shows that the addition of proposed stylistic features to the baseline features, results in improvement of classification accuracies across the four datasets. The improved performance due to stylistic features at individual levels indicates their ability to capture new notions of style and the significant increase when \textit{all} the style elements are used together, reinforces the need of a multi-level stylistic analysis. As a sidenote, in Table \ref{tab:results_AA_Ablation}, it can observed that addition of surface-level stylistic elements does not improve the classification accuracy as much as addition of lexical and syntactic elements do. This can be attributed to the fact that most of the existing style-based features in the baseline can be identified as surface-level features, whereas very few can be identified as lexical or syntactic.

\section{Emotion Prediction}
\label{sec:emotionPrediction}

We now illustrate the value of a multi-level analysis of style in solving the task of \textit{fine-grained} emotion classification. While emotion can be classified on a discrete level (e.g., happy, sad, excited, etc.) we focus on a fine-grained classification using valence, arousal, and dominance values \cite{strapparava2007semeval,buechel2017emobank}.
The manner in which meaning is conveyed influences the emotion it evokes in readers of a given text \cite{wise2009words,kao2012computational}. While the relationship between content and emotion has been studied extensively \cite{subasic2001affect,neviarouskaya2011affect,kantrowitz2003method}, owing to availability of language-specific resources \cite{mohammad2013nrc,mohammad2018obtaining}, little research has been done to study the relationship between style and emotion \cite{kao2012computational}. The motivation for considering the task of emotion prediction in this work is twofold: \textit{(a)} analyze the role of stylistic aspects of text in predicting emotion, and \textit{(b)} analyze the value of having a multi-level stylistic representation as proposed in this work. 

We consider the method proposed by \citeauthor{akhtar2018multi} (\citeyear{akhtar2018multi}) as a baseline and concatenate our proposed multi-level stylistic features with their existing features. The baseline and the proposed modification is evaluated on two standard datasets for emotion classfication: the EmoBank dataset \cite{buechel2017emobank} and the Facebook posts dataset \cite{preoctiuc2016modelling}.
The EmoBank dataset comprises of $10,062$ tweets across multiple domains (e.g. blogs,
news headlines, fiction etc.). Each tweet has three scores representing valence, arousal and dominance of emotion on a continuous range of 1 to 5. The Facebook posts dataset contains $2,895$ social media posts that are annotated on a nine-point score with valence and arousal scores by two psychologically trained annotators. To ensure consistency while comparing results with the baselines, for experiments, we adopt 70-10-20 split
for training, validation and testing, respectively. As stated in the work by \citeauthor{akhtar2018multi} (\citeyear{akhtar2018multi}), we perform 10-fold cross-validation for the evaluation. We also use the same training and evaluation setup, along with same model hyperparameters, to ensure meaningful comparisons. For more implementations details, please refer to \citeauthor{akhtar2018multi} (\citeyear{akhtar2018multi}).
 
\subsubsection{Baselines:} \citeauthor{akhtar2018multi} (\citeyear{akhtar2018multi}) propose a multi-task ensemble that combines the learned representations of three independently trained deep learning models (i.e., a Convolutional Neural Network (CNN), a Long
Short Term Memory (LSTM), and a  Gated Recurrent
Unit (GRU) network) and a hand-crafted feature vector that comprises of features like word and character tf-idf, lexicon-based sentiment scores, count of positive and negative words, etc. The multi-task ensemble is essentially a multi-layer perceptron (MLP) with two shared hidden layers and two task-specific hidden layers that cater to the specific need of individual tasks. Their motivation of solving the three regression problems (one for each valence, arousal and dominance) in a multi-task setup arises from the intuition that these related tasks can help the joint-model learn effectively from shared representations while achieving better generalization. To assess the role of our proposed stylistic features in predicting emotion, we concatenate them with the handcrafted features of \citeauthor{akhtar2018multi} For comparison with other existing state-of-the-art methods, we also include the performance of the System proposed by \citeauthor{preoctiuc2016modelling} (\citeyear{preoctiuc2016modelling}).
 
As it can be observed from results presented in Table \ref{tab:results_EP_SoTA}, the addition of new multi-level stylistic features leads to significant improvement in the Pearson correlation coefficient, over multiple baselines.  Pearson correlation coefficient measures the linear correlation between the actual and predicted scores and has been used extensively in prior art \cite{mohammad2017wassa,preoctiuc2016modelling}.

In Table \ref{tab:results_EP_SoTA}, we quantify the improvement brought by incorporating stylistic elements at individual levels. The average change in Pearson coefficient correlation, over $20$ different experimental runs, is reported in Table \ref{tab:results_EP_SoTA} along with the statistical significance (\textit{t-test}) of reported results. As it was the case with the task of author attribution (see Table \ref{tab:results_AA_Ablation}), inclusion of \textit{all} multi-level stylistic features allows the model to capture new notions of style and reinforces the need of a multi-level stylistic analysis. Additionally, referring back to our motivation of choosing the emotion prediction task, we provide empirical evidence that stylistic aspects do correlate with valence, arousal and dominance values. The questions around causal \textit{significance} and \textit{extent} \cite{pearl2010causal,wang2018blessings} of stylistic aspects towards evoked emotion are yet to be answered, and are left as a part of future work. However, we expect the interpretability of the proposed stylistic features to aid in establishing causal relationships. 

\section{Discussion of Results}
\label{sec:discussion}
To illustrate the value of the proposed multi-level representation of style, we focused on three tasks: authors' style analysis, authorship attribution and emotion prediction. Given this, it becomes essential to emphasize that this work does not aim to propose novel approaches to any of the aforementioned tasks. The primary aim of the work is an effort to establish a structured multi-level understanding of style in text that can facilitate in better modeling of style. We substantiate the value of such an understanding by giving empirical evidences in Table \ref{tab:results}, \ref{tab:results_AA_SoTA}, \ref{tab:results_AA_Ablation} and \ref{tab:results_EP_SoTA}.

To summarize the empirical evidences, by quantifying and analyzing the writing style of $5$ English authors, we demonstrate that the proposed stylistic features provide interpretable and coherent insights about an author's style. When the proposed multi-level stylistic features are added in a simplistic way to solve the tasks of authorship attribution and emotion prediction, they further improve the performance of existing state-of-the-art approaches. Specific to the task of emotion prediction, we also demonstrate that stylistic aspects of text have a correlation with the emotion it evokes in its readers. 

An interesting aspect that is highlighted by solving the tasks of authorship attribution and emotion prediction is the varying extent to which stylistic elements at different levels contribute towards solving the task. For instance, in Table \ref{tab:results_EP_SoTA}, the lexical and syntactic-level elements of style add significantly more value to the task of emotion prediction than surface-level elements. This claim is substantiated as we note that the original handcrafted features used in the baseline are devoid of any stylistic features whatsoever. The proposed structure provides more holistic interpretability while modeling style to solve related tasks.

\section{Related Work}
\label{sec:relatedWork}
In this section we provide a comprehensive description of prior related work. We start by discussing in detail the work that aims to computationally model style in text. This is followed by a brief discussion of existing approaches to analyze the style of authors and solve the task of authorship attribution and emotion prediction.  

\subsubsection{Understanding Style in Text:}
As mentioned earlier, while there is a recent focus on solving style-related natural language tasks \cite{li2018delete,prabhumoye2018style,jhamtani2017shakespearizing,shen2017style}, there has been a decline in efforts that aim to identify what style constitutes and provide a holistic task-independent understanding of it \cite{tikhonov2018wrong,crystal2016investigating}. Given the recent advancements in machine learning and data-driven approaches in style-related problems, it is imperative that we look back to align our understanding of style to work with and aid recent methods.

Efforts to understand style range back to the work by \citeauthor{dimarcoStyle} (\citeyear{dimarcoStyle}) where they provide a grammar of style while translating realizations from one language to the other. To do so, they introduce the notion of \textit{internal} \textit{stylistics} of source and target languages and a method to map these internal notions of style. The internal stylistics of a language relate to its linguistic characteristics and capture aspects like abstraction, dynamism, clarity, and formality. However, given the recent standing of the larger field, it is difficult to leverage these linguistic-based internal stylistics to aid data-driven computational models. Similar problems arise with other prior works \cite{brewer1984reconstructive,semino2002cognitive,freeborn1996style} where the extension of linguistic intuitions to currently prevalent modeling approaches is non-trivial.

More recently, there have been efforts to quantify stylistic features to enable style-based text categorization \cite{koppel2003corpus,argamon1998style}. Since text categorization itself encompasses several downstream tasks (e.g., sentiment analysis, genre classification, authorship attribution, etc.) there is a tendency to define style -- and consequently, the stylistic features -- in a task-specific manner \cite{tikhonov2018wrong}.  It is also notable that the definition of stylistic features are inconsistent among prior works. For instance, \citeauthor{brooke2013multi} (\citeyear{brooke2013hybrid,brooke2013multi}) focus primarily on lexical-level stylistic aspects while \citeauthor{feng2012syntactic} (\citeyear{feng2012syntactic}) focus on aspects of style at syntactic-level.

Our proposed work adds to the current research by providing a structured understanding of stylistic elements at surface, lexical, syntactic and semantic-level. We also demonstrate that such an understanding, which is rooted in linguistically rich intuitions, can be used to obtain a multi-level representation of style which can be further used to improve the performance of state-of-the-art data-driven approaches across multiple tasks. 

Next, we discuss the prior works that aim to solve the aforementioned tasks by taking stylistic aspects of text into account.

\subsubsection{Authors' Style Analysis:}
There are several works that aim to quantify stylistic linguistic intuitions to analyze the writing style of well known authors \cite{mccarthy2006analyzing,peng2002quantitative,forgeard2008linguistic}. They leverage features ranging from cohesion measures\footnote{Cohesion is the grammatical and lexical linking within a text or sentence that holds a text together and gives it meaning. For instance, the linking that enables us to understand the reply in the following conversation: \textit{(A)} ``Where are you going?" \textit{(B)} ``To dance."} to reading difficulty. They often demonstrate the value of their identified features by relating qualitative insights with quantified representations. Taking motivation from here, we demonstrate the value of our proposed stylistic representation by building a coherent understanding (across multiple levels) of writing style of $5$ well known English authors. 

\subsubsection{Authorship Attribution:}
Recent approaches to the task of authorship attribution make use of a mix of content and style-based features \cite{sari2018topic}. While the content-based features are majorly character or word-based n-grams \cite{sari2017continuous,kevselj2003n}, the style-based features include shallow features like function word frequencies, count of \textit{hapax} \textit{legomena}, etc., and deep linguistic features like context free production frequencies and semantic relationship frequencies \cite{gamon2004linguistic}. Addition of our proposed stylistic representation facilitates modeling of newer notions of style in a structured and interpretable manner and improves the performance of existing state-of-the-art models. 

\subsubsection{Emotion Prediction:}
Features like count of positive and negative words \cite{wiebe2006word}, count of words  matching each emotion from the NRC Word-Emotion Association Lexicon \cite{mohammad2013crowdsourcing}, word and character tf-idf have been extensively used for predicting emotion \cite{akhtar2018multi,preoctiuc2016modelling}. However, the efforts that aim to analyze the influence of stylistic aspects on emotion have been scarce \cite{kao2012computational}. We demonstrate that linguistically motivated stylistic features are not only correlated with emotion, but also help in improving the performance of existing emotion prediction approaches.

\section{Conclusion and Emerging Directions}
\label{sec:conclusion}
Understanding style is an important aspect of modeling inherent subjectivity in text. We presented a linguistically-motivated process to understand and qualify stylistic aspects of text 
at lexical, syntactic, and semantic-level. Using existing methods to quantify these style-related linguistic intuitions, we analyzed the writing style of $5$ authors, and solved the task of authorship attribution and emotion prediction. We demonstrated that the style of an author is a compound factor of various stylistic elements, validating the need of a multi-level analysis of style. We also improve the performance of existing state-of-the-art approaches to authorship attribution and emotion prediction task by modeling style in a structured and interpretable manner. To strengthen the empirical evidences, we conducted experiments on datasets that contained text from diverse topics and domains (social media posts, literary texts, legal documents and movie reviews).

Such a multi-level style analysis, by being able to incorporate broader notions of style in a more structured and interpretable manner, can aid in understanding the causal influence of style towards psycholinguistic concepts like formality, sentimentality, politeness, etc., by deconfounding other potential causes like topic. In future, we aim to analyze the influence of style on psycholinguistic concepts in further detail. 

\balance
\bibliographystyle{aaai}
\bibliography{biblio}
\end{document}